\RequirePackage{amsmath}
\documentclass[runningheads,hidelinks]{llncs}
\usepackage[T1]{fontenc}
\usepackage{graphicx}
\usepackage{color}
\usepackage[colorlinks = true,
            linkcolor = black,
            urlcolor  = blue,
            citecolor = black,
            anchorcolor = black]{hyperref}

\usepackage{multirow}
\usepackage[table]{xcolor}

\usepackage{booktabs} 

\usepackage{mathtools}

\usepackage[capitalize,noabbrev]{cleveref}
\crefname{section}{Sec.}{Secs.}
\Crefname{section}{Section}{Sections}
\Crefname{table}{Table}{Tables}
\crefname{table}{Table}{Tables}
\Crefname{equation}{Equation}{Equations}
\crefname{equation}{Eq.}{Eqs.}
\Crefname{figure}{Figure}{Figures}
\crefname{figure}{Fig.}{Figs.}

\newcommand{\cifar}{\textsc{Cifar10}}
\newcommand{\imagenetonehunderd}{\textsc{ImageNet100}}
\newcommand{\imagenettwohundred}{\textsc{ImageNet200}}
\newcommand{\imagenet}{\textsc{ImageNet}}
\newcommand{\imagenetc}{\textsc{ImageNet-C}}

\emergencystretch=\maxdimen
\begin{document}
%
\title{PushPull-Net: Inhibition-driven ResNet robust to image corruptions}
\titlerunning{PushPull-Net: Inhibition-driven ResNet robust to image corruptions}
%
\author{Guru Swaroop Bennabhaktula\inst{1}\orcidID{0000-0002-8434-9271} \and
Enrique Alegre\inst{2}\orcidID{0000-0003-2081-774X} \and
Nicola Strisciuglio\inst{3}\orcidID{0000-0002-7478-3509} \and
George Azzopardi\inst{1}\orcidID{0000-0001-6552-2596}}
\authorrunning{G.S. Bennabhaktula et al.}
%
\institute{University of Groningen, the Netherlands
\email{\{g.s.bennabhaktula,g.azzopardi\}@rug.nl} \and
University of Léon, Spain
\email{enrique.alegre@unileon.es} \and
University of Twente, the Netherlands
\email{n.strisciuglio@utwente.nl}
}
\maketitle              
\begin{abstract}
We introduce a novel computational unit, termed PushPull-Conv, in the first layer of a ResNet architecture, inspired by the anti-phase inhibition phenomenon observed in the primary visual cortex. This unit redefines the traditional convolutional layer by implementing a pair of complementary filters: a trainable push kernel and its counterpart, the pull kernel. The push kernel (analogous to traditional convolution) learns to respond to specific stimuli, while the pull kernel reacts to the same stimuli but of opposite contrast. This configuration enhances stimulus selectivity and effectively inhibits response in regions lacking preferred stimuli. This effect is attributed to the push and pull kernels, which produce responses of comparable magnitude in such regions, thereby neutralizing each other. The incorporation of the PushPull-Conv into ResNets significantly increases their robustness to image corruption. Our experiments with benchmark corruption datasets show that the PushPull-Conv can be combined with other data augmentation techniques to further improve model robustness. We set a new robustness benchmark on ResNet50 achieving an $mCE$ of 49.95\% on \imagenetc~when combining PRIME augmentation with PushPull inhibition.

\end{abstract}

\section{Introduction}
\label{sec: intro}

One of the key challenges facing ConvNets is their ability to generalize to out-of-distribution images~\cite{wang2024surveyrobustnesscomputervision}. This distribution shift may manifest itself in several forms, such as image corruption~\cite{hendrycks2019benchmarking}, stylized image transformations~\cite{karras2020analyzing}, rendition~\cite{hendrycks2021many}, and adversarial samples~\cite{shiva2017simple}, among others. Image corruption is frequently encountered and can happen without any deliberate interference or adversary involvement. Corruptions can be of several types. This work focuses on improving the robustness of ResNet to common image corruptions such as noise, blur, weather, and digital.

\begin{figure}[t]
  \centering
   \includegraphics[width=0.4\linewidth]{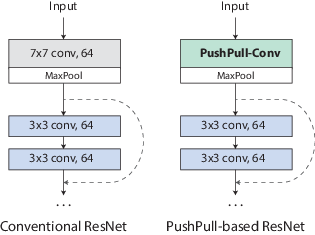}
   \caption{The proposed approach involves substituting the first convolutional layer with PushPull-Conv, enhancing robustness with minimal computational overhead.}
   \label{fig:push_pull_based_resnet}
\end{figure}

Among a few strategies to tackle this problem~\cite{wang2023robustness}, a couple of popular ones include (a) data augmentation in the training phase and (b) by design, where intrinsic changes are made to the model's architecture. Regularization strategies like dropout, batch normalization, and weight decay enhance model generalization and reduce overfitting but help little to improve model robustness against image corruption.

Data augmentation has considerably improved model robustness~\cite{hendrycks2019benchmarking,hendrycks2021many,yin2019fourier,Vaish_2024_CVPR}. However, mere data augmentation cannot address this problem sufficiently~\cite{yin2019fourier}. Improvement in the architectural design of ConvNets along with data augmentation strategies have improved model robustness~\cite{liu2021group}. We contribute to enhancing ConvNet architectural design for improved model robustness. While~\cite{liu2021group} deals with inhibiting the most significant activations, the proposed push-pull inhibition takes a contrasting approach and reduces signal responses in regions lacking preferred stimuli.

ConvNets trained on \imagenet~\cite{ILSVRC15} are biased towards classifying texture over an object's shape~\cite{geirhos2018imagenet}. This is in contrast to the behaviour of humans, and Geirhos~et~al.~\cite{geirhos2018imagenet} showed that introducing shape bias improves the model's robustness and accuracy on \imagenet. Unlike~\cite{geirhos2018imagenet}, we introduce such an ability using intrinsic changes to the ConvNet rather than by stylized data transforms. Our methodology is biologically inspired by the work of Hubel and Wiesel~\cite{hubel1962receptive,hubel1979brain} on simple cells in the primary visual cortex of mammals and by the antiphase or push-pull inhibition that they exhibit~\cite{borg1998visual}. We propose a novel inhibition-driven computational unit inspired by the behaviour of these simple cells, employing a push-pull based inhibition mechanism as detailed below.

The proposed computational unit, PushPull-Conv, consists of excitatory and inhibitory convolutions, in other words, push and pull convolutions. The pull kernel is the complement of the push kernel.
The PushPull-Conv unit, with its simultaneous use of push (excitatory) and pull (inhibitory) filters, offers a sophisticated approach to visual processing that improves selectivity. In the presence of a preferred stimulus, the PushPull-Conv unit exhibits a pronounced output due to the strong response from the push filter and a weak response from its complementary pull filter, enhancing the detection of specific features. In contrast, applying a low-pass filter first would attenuate high-frequency details, potentially diminishing critical information before the push filter acts, thereby affecting its effectiveness in identifying the preferred stimulus. Additionally, in the absence of the preferred stimulus, the PushPull-Conv unit ensures minimal output by balancing the similar responses of the push and pull filters, thereby cancelling each other out and enhancing overall selectivity. This mechanism contrasts with the low-pass-then-push approach, which might reduce high-frequency noise but could amplify non-preferred low-frequency stimuli, leading to less specificity. Furthermore, the PushPull-Conv unit dynamically adjusts its response across the entire spectrum of spatial frequencies, providing a high degree of selectivity and sensitivity, while the sequential filtering approach lacks this nuanced control and adaptability, making it less effective in discriminating between preferred and non-preferred stimuli with the same level of precision.

The main contribution of this work is three-fold.
(1) A new computational model for push-pull inhibition is proposed. We call this PushPull-Conv and propose to replace the traditional convolutional unit as the first layer of the ConvNet. This innovative architectural change not only heightens precision in identifying preferred stimuli but also effectively attenuates non-preferred stimuli across all frequency ranges. This introduces a robust alternative to traditional approaches that sequentially apply noise-specific filters followed by push-only filtering, thereby offering a more balanced and comprehensive method for feature detection and image processing. (2) Data augmentation strategies such as AugMix, AutoAug, TrivialAugment, and PRIME are known to improve model robustness. Our research shows that combining PushPull-Conv with these data augmentation techniques further enhances model robustness, yielding additional improvements. (3) Propose a new metric to quantify the trade-off between robustness and accuracy. 

\section{Related Works}
\label{sec: related_work}

Deep networks can be susceptible to image perturbations that humans cannot perceive~\cite{recht2018cifar,carlini2017towards}, mainly because the model relies on features unrelated to the task. For instance, networks (such as ResNet) trained on \imagenet~for object detection are biased towards texture-based features instead of shape-based, when in fact, humans rely more on the shape for image classification~\cite{geirhos2018imagenet}. Although adversarial robustness is not within the scope of this study, it is worth mentioning that this research direction has seen tremendous progress with every adversarial attack~\cite{carlini2016defensive,carlini2017adversarial} countered with adversarial defences~\cite{papernot2016distillation,metzen2017detecting}, and vice-versa. Some other methods that attempt to improve robustness are model pre-training~\cite{hendrycks2019using}, enhancing performance on in-distribution dataset~\cite{taori2019robustness}, and using larger models~\cite{hendrycks2019benchmarking}.


Data augmentation improves the generalization and robustness of deep models. For instance, image transformations such as flips, crops, resize, affine transforms and other methods are commonly used to augment the training data. A few popular techniques include, cut-mix~\cite{yun2019cutmix}, AutoAug~\cite{Cubuk_2019_CVPR}, AugMix~\cite{hendrycks2019augmix}, PRIME~\cite{PRIME2022}, TrivialAugment~\cite{Muller_2021_ICCV}, DeepAugment~\cite{hendrycks2021many}, and adversarial training~\cite{goodfellow2014explaining,wang2019bilateral}, among others. 
Hendrycks~et~al.~\cite{hendrycks2021many} found that using larger models and artificial data augmentation improves model robustness across several corruption types. Interestingly, they showed that out-of-distribution training using the intermediate feature maps from deep networks (artificial data) also improves model robustness to natural image corruptions. Yin~et~al.~\cite{yin2019fourier} performed an analysis of model robustness in the Fourier domain. They showed how models trained with certain augmentations, such as Gaussian noise, and adversarial training improve model robustness against several noise and blur corruptions while reducing robustness to fog and contrast corruptions. Though data augmentation can address model robustness to a certain extent, it can be made effective only with the availability of a lot of data \cite{schmidt2018adversarially}.  

An orthogonal line of research is to make improvements to the model architecture~\cite{vasconcelos2020effective,liu2021group,strisciuglio2020enhanced}. Such methods could be used in conjunction with data augmentation to further enhance model robustness. For instance, Liu~et~al.~\cite{liu2021group} proposed a regularization technique called group inhibition, to extract auxiliary features for robust classification. When combined with data augmentation,~\cite{vasconcelos2020effective,liu2021group} show that their methods improve results in comparison to when only data augmentation is used. 
Our work also uses a similar strategy to make the models more robust by introducing changes to the architecture. However, our work differs in the specifics of the architectural design. In~\cite{vasconcelos2020effective,zhang2019making}, it was proposed to include a strided blur filter following strided convolutions to address the aliasing problem in ConvNets. These modifications were done to several ConvNet layers. In contrast, we propose to replace only the first layer of a ConvNet with the proposed PushPull-Conv layer, resulting in a more computationally efficient solution.

Our architectural design is biologically inspired by the push-pull phenomenon exhibited by simple cells in the mammalian brain~\cite{hubel1962receptive,hubel1979brain,ferster1988spatially,borg1998visual,hirsch1998synaptic}. In~\cite{azzopardi2014push}, a computational model of this phenomenon was proposed. The resulting model can be used as a contour detection operator that achieves superior robustness against corruption. It was later embedded in the RUSTICO line detector~\cite{strisciuglio2019robust} and coupled with surround inhibition~\cite{melotti2020robust}.
Taking inspiration from simple cells that receive such inhibition, we propose a unit, namely the PushPull Convolutional unit or PushPull-Conv for short. The design of this unit is further elucidated in \cref{sec: The Push Pull Unit}.

While our work might appear akin to that of~\cite{strisciuglio2020enhanced} at first glance, it fundamentally diverges in both methodology and scope. 
Firstly, we propose a modified design of the push-pull computational unit and make it configurable with learnable inhibition strengths. In the earlier attempt, with learnable inhibition the robustness became inconsistent, slowed down training, and was left for future work. Trainable inhibition in our approach leads to consistent results without slowing down the training and is a default choice for hyperparameters.
Secondly, the push-pull approach in~\cite{strisciuglio2020enhanced} is merely an \textit{erosion operation}. Without the upsampling factor in~\cite{strisciuglio2020enhanced}, the pull component would be redundant. Our new way of determining the pull kernel results in a push-pull component that is effective even without pooling. Thirdly, we show that our unit complements other data augmentation methods.

\section{Approach}

\subsection{The Push Pull unit}
\label{sec: The Push Pull Unit}
The proposed computational design for performing push-pull convolutions is the rectified difference between the responses of the push and the pull convolutions on the input image (\cref{fig:pushpull_unit}). The push kernel is analogous to the traditional convolutional unit in ConvNets and is defined as:

\begin{figure}[t]
  \centering
   \includegraphics[width=0.6\linewidth]{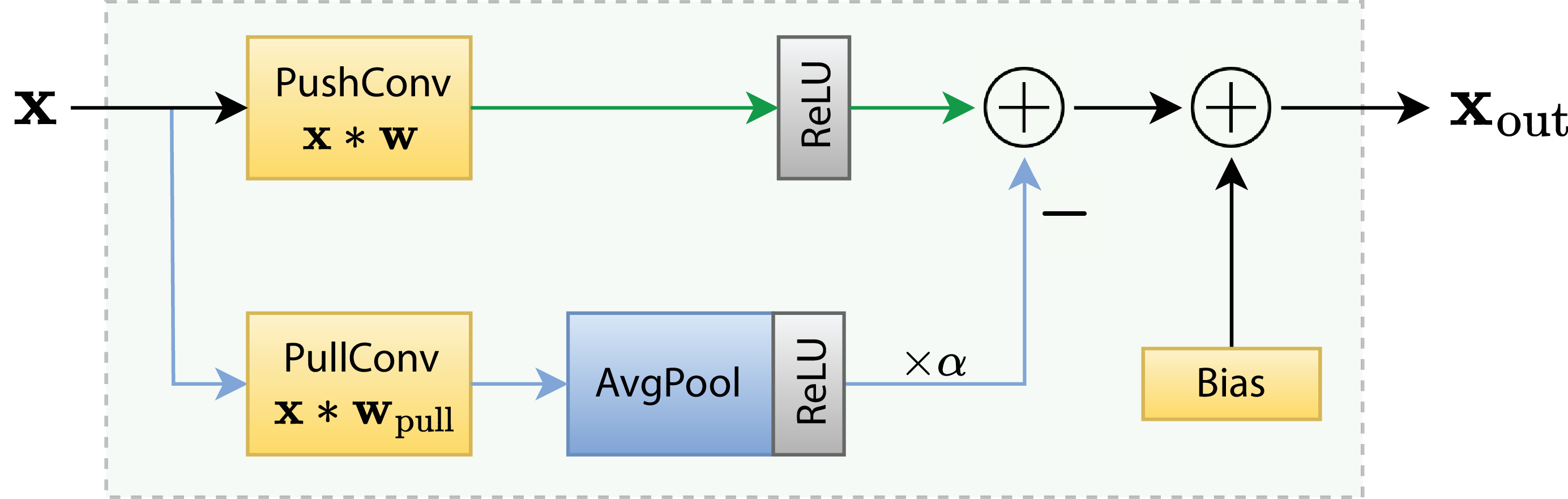}
   \caption{The proposed push-pull computation unit. Refer \cref{eq: pupu_unit}.}
   \label{fig:pushpull_unit}
\end{figure}

\begin{equation}
    \mathbf{x}_{\text{out}} = \mathbf{x}*\mathbf{w} + b
\end{equation}
\noindent where $\mathbf{x}$ denotes the input signal, $\mathbf{w}$ denotes the convolutional kernel, and $b$ is the bias. The pull kernel is derived from the push kernel by taking its complement such that it responds to the same preferred patterns of the push kernel but of opposite contrast. This is inspired by the receptive field with the antiphase of simple neurons that experience this kind of inhibition~\cite{hubel1979brain}. Formally, we determine the pull kernel as follows.

\begin{equation}
\label{eq: non_scaled_pull_kernel}
    \mathbf{w}_{\text{pull}}^{'} = 1 - \frac{\mathbf{w} - {w}_{\text{min}}}{{w}_{\text{max}} - {w}_{\text{min}}}
\end{equation}

\noindent where $\mathbf{w}$ denotes the weights of the push kernel and its maximum and minimum values are represented by ${w}_{\text{max}}$ and ${w}_{\text{min}}$, respectively. The values of $\mathbf{w}_{\text{pull}}^{'}$ are rescaled to match that of $\mathbf{w}$, which determines the final pull kernel:

\begin{align}
\label{eq: scaled_pull_kernel}
    \mathbf{w}_{\text{pull}} &= \mathbf{w}_{\text{pull}}^{'} \times ({w}_{\text{max}} - {w}_{\text{min}}) + {w}_{\text{min}} \\
\label{eq: scaled_pull_kernel_simplified}   
    \mathbf{w}_{\text{pull}} &= -\mathbf{w} + {w}_{\text{max}} + {w}_{\text{min}}
\end{align}

\begin{figure}[t]
  \centering
   \includegraphics[width=0.5\linewidth]{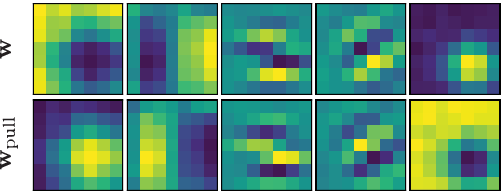}
   \caption{(Top row) Five randomly selected filters from the conv1 layer of ResNet50 (trained on \imagenet). For a visual illustration, only the first filter channel is depicted. (Bottom row) The corresponding pull kernels as determined by \cref{eq: scaled_pull_kernel_simplified}.}
   \label{fig:resnet50_push_pull_kernels}
\end{figure}

\Cref{eq: scaled_pull_kernel} can be simplified to \cref{eq: scaled_pull_kernel_simplified} using \cref{eq: non_scaled_pull_kernel}. \Cref{fig:resnet50_push_pull_kernels} shows a few examples of push and pull kernels.

Since the pull kernel is a non-linear function of the push kernel, it indirectly introduces non-linearity into the pull response. The pull response can be considered as inhibition and it is subtracted from the response generated by the push kernel. This results in the push-pull computation:

\begin{equation}
\label{eq: pupu_unit}
    \mathbf{x}_{\text{out}} = r(\mathbf{x}*\mathbf{w}) - \alpha \cdot r(\mathbf{x}*\mathbf{w}_{\text{pull}}*\mathbf{1}) + b
\end{equation}

\noindent where $\alpha$ is the pull inhibition factor, $r$ is the ReLU activation, and $\mathbf{1}$ represents the average filter. 
The average pooling increases the receptive field size of the pull component. It is incorporated in our work because of the great benefit it contributed to the computational model as first proposed in~\cite{azzopardi2014push}. It is inspired by the neurophysiological discovery that interneurons have broader receptive fields than those of the neurons they inhibit~\cite{martinez2005receptive}. We extend this idea to any arbitrary (push) kernel learnt by ConvNet, and through push-pull computation to possibly achieve the properties of a band-pass filter. The factor $\alpha$ controls the degree of pull inhibition and can either be configured as a trainable parameter or set to a fixed value, thereby making it a hyper-parameter. We experimentally show that the latter scheme is useful when manually controlling the trade-off between the model performance for in- and out-of-distribution data (refer to \cref{sec: tradeoff_metric} for further details).   

\Cref{fig:pushpull_unit} depicts the proposed push-pull unit. Note that the bias term is only added at the final stage. Both the push and pull convolutions are performed without any bias. This is done to simplify the push-pull design and to remove any undesired influence of the bias when the pull response is subtracted from the push.

\subsection{Characteristics of Push Pull convolutions}

To illustrate the design of the push-pull kernel, let's consider a simpler form of \cref{eq: pupu_unit} without the non-linear activation $r$. In this case, \cref{eq: pupu_unit} simplifies to:

\begin{align}
    \mathbf{x}_{\text{out}} &= \mathbf{x}*\mathbf{w} - \alpha \cdot (\mathbf{x}*\mathbf{w}_{\text{pull}}*\mathbf{1}) + b \\
    &= \mathbf{x}*(\mathbf{w} - \alpha \cdot \mathbf{w}_{\text{pull}}*\mathbf{1}) + b \\
    \label{eq: pupu unit without non-linearities}
    &= \mathbf{x}*f(\mathbf{w}; \alpha, \mathbf{1}) + b 
\end{align}

\noindent where the push-pull kernel is represented by the non-linear function $f(\mathbf{w}; \alpha, \mathbf{1})$.

\begin{figure}
  \centering
   \includegraphics[width=0.7\linewidth]{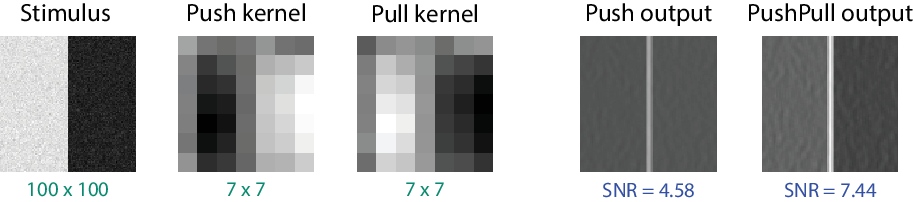}
   \caption{Illustration of a push-pull filtering with a simulated input corrupted by Gaussian noise. The push kernel is learned by ResNet50. 
   SNR computed as $20\log_{10}(A_{s}/A_n)$, where $A_s$ is the average across a 5-pixel vertical edge, and $A_n$ is the average response in the background.}
   \label{fig:pushpull_simulated_example}
\end{figure}

\begin{figure*}
  \centering
   \includegraphics[width=1.0\linewidth]{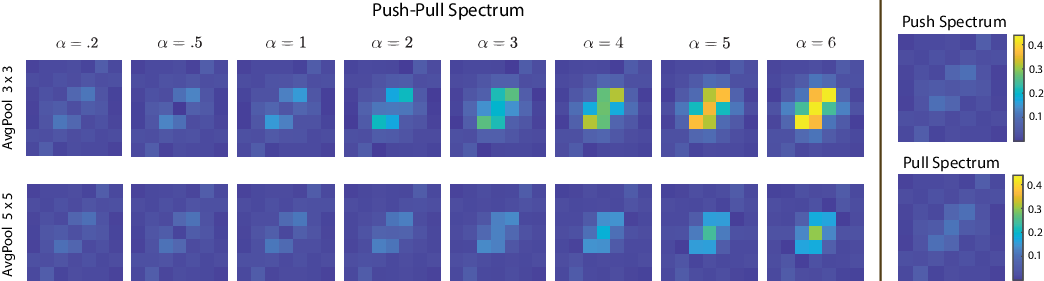}
   \caption{Fourier spectral analysis of push-pull filters in ResNet50. The averaged push spectrum of the 64 push kernels is presented in the top right, while the corresponding pull spectrum is shown at the bottom right. These push and pull spectrum are then combined with various choices of hyperparameters $\alpha$ and AvgPool, and the resulting push-pull spectra are displayed on the left. For visual illustration, only the first filter channel is depicted.}

   \label{fig: fourier_analysis}
\end{figure*}

We conducted an analysis in the Fourier domain to gain deeper insights into the characteristics of push-pull convolutions, by considering all the 64 kernels from the first layer of a ResNet50 trained on \imagenet. The computed spectra are presented in \cref{fig: fourier_analysis}. On a careful examination, it can be seen that for smaller values of inhibition strength $\alpha$, the push-pull filters exhibit the property of band-pass. This trend is observed for both configurations of pooling. Note that, band-pass filtering has been used for edge detection, for instance, using Difference-of-Gaussians (DoG) filters. Therefore, for smaller values of $\alpha$, the push-pull mechanism induces a minor bias for extracting shape-based features. Thus, it can assist the model in becoming more robust to high-frequency corruptions. We illustrate this behaviour further with a simple example in \cref{fig:pushpull_simulated_example}, where the SNR improves with the push-pull mechanism. 

\subsection{Embedding PushPull-Conv in ConvNets}
\label{sec: Embedding PushPull-Conv in ConvNets}
The idea is conceptually simple and easy to implement. We replace the first convolutional layer in a ConvNet with PushPull-Conv, as shown in \cref{fig:push_pull_based_resnet}. 
The rationale behind using such a computational unit in the first layer of the network is also biologically inspired. Neurophysiology studies show that this kind of inhibition is found in the early stages of the visual system. Besides simple cells in the primary visual cortex, the lateral geniculate nucleus (the intermediate unit between the retina and the primary visual cortex) contains cells with centre-surround receptive fields. These receptive fields can be modelled by the Laplacian of Gaussian or DoG operators. DoG filters can also be considered a type of push-pull filter, and they achieve band-limited selectivity. As shown in \cref{fig: fourier_analysis}, with appropriately chosen inhibition strength $\alpha$, the filters in the PushPull-Conv layer tend to act as band-pass filters and potentially suppress high-frequency corruptions, thereby, correcting the distributional shift for the layers following it in the network.  

\section{Experiments and Results}

To evaluate the effectiveness of our proposed approach on out-of-distribution images, we train our models on \cifar~\cite{krizhevsky2009learning} and \imagenet~\cite{ILSVRC15} datasets, and test the resulting models on \cifar-C and \imagenetc~\cite{hendrycks2019benchmarking}. The latter two datasets contain the corrupted versions of images in the former two datasets. The test images are perturbed with 15 types of image corruptions (broadly categorized into noise, blur, weather, and digital) and 5 levels of severity for each corruption type (see Fig.~9 in the \href{https://github.com/bgswaroop/pushpull-conv}{Supplementary Material}).

To analyze how the robustness of the PushPull-Conv-based model scales with increasing dataset size, two additional subsets of \imagenet~were created with 100 and 200 classes each (\imagenetonehunderd~and \imagenettwohundred). The classes were randomly chosen and their corresponding corrupted test set versions are referred to as \textsc{ImageNet100-C} and \textsc{ImageNet200-C}.


\subsection{Evaluation metrics}

We use the mean corrupted error as a metric to measure robustness~\cite{hendrycks2019benchmarking}. Let the error made by a classifier $f$, on a corruption type $c$, with a severity level $s$, be denoted by $E^{f}_{c,s}$. Then the (absolute) mean corruption error for a classifier $f$ is defined as:
\begin{equation}
\label{eq: mCE}
    mCE = \frac{1}{|C|} \sum_{c \in C}\biggl({CE}^{f}_{c}\biggr) = \frac{1}{|C|} \sum_{c \in C}\biggl({\frac{1}{S} \sum_{s=1}^{S}{E^{f}_{c,s}}} \biggr)
\end{equation}
\noindent where, $mCE$ is the mean of corruption errors $CE_{c}^{f}$ over the set of corruptions $C \in \{\text{Gussian Noise, Shot Noise, ..., JPEG}\}$. When comparing the performance of a model $f$ to a baseline $b$, it is useful to represent it as the mean of relative corruption errors, which is given by:
\begin{equation}
\label{eq: mrCE}
    mrCE = \frac{1}{|C|} \sum_{c \in C}\biggl({CE}^{f}_{c} / {CE}^{b}_{c} \biggr)
\end{equation}

\subsection{Experimental setup}

We investigate the model robustness by replacing the first layer of ResNet with the proposed PushPull-Conv. As the original ResNet architecture for \imagenet~had a conv1 layer with 64 filters of size $7\times7$ and a stride of 2, we retain the same configuration for the PushPull-Conv and explore various settings for the hyperparameters - \textit{average kernel size} and the \textit{inhibition strength}. 
As the $mrCE$ scores (refer to \cref{eq: mrCE}) are always computed with respect to a baseline, we hereby always refer to the vanilla ResNet as the baseline. With this setting, we evaluate the $mrCE$ scores of the PushPull variants keeping the corresponding vanilla ResNet as the baseline. For the experiments on the \cifar~dataset, we modify the conv1 layer to have 64 filters with size $3\times3$ and with a stride of 1. This change in hyperparameters was made to handle the smaller image dimensions ($32\times32$ pixels) of that dataset.

The classification models were optimized using the Stochastic Gradient Descent (SGD) with a momentum of 0.9 and a weight decay of 1e-5. A cyclic learning rate schedule~\cite{smith2019super} was employed wherein the learning rate was increased from 0.05 to 1.0 for the first 30 per cent of the steps and decayed thereafter to 5e-5. The learning rate warm-up and decay both followed a cosine policy. The models were trained for 20 epochs with a batch size of 128. The objective function was determined using categorical cross-entropy loss. This training recipe was chosen to enable faster experimentation and validate our hypothesis. The aforementioned recipe was used to train ResNet18 and ResNet50 on \imagenet. For hyperparameters concerning other models and dataset combinations, we refer the reader to our source code\footnote{\url{https://github.com/bgswaroop/pushpull-conv}}.

\subsection{Experiments}
Experiments were conducted using ResNet18 and ResNet50 models, on 4 different datasets \cifar, \imagenetonehunderd, \imagenettwohundred, and \imagenet. 

\begin{table*}[t]
  \caption{Classification results. D denotes the dataset, $E$ is the top-1 error rate on the clean test set. The corruption types are \textbf{Noise} - (Ga)ussian, (Sh)ot, and (Im)pulse, \textbf{Blur} - (De)foucus, (Gl)ass, (Mo)tion, and (Zo)om, \textbf{Weather} - (Sn)ow, (Fr)ost, and (Fo)g, and \textbf{Digital} - (Br)ightness, (Co)ntrast, (El)astic, (Pi)xelate, and (Jp)eg. The push-pull variants are shown underneath the respective baselines. The baselines are marked with {\scalebox{1.0}{\textcolor{teal}{(b\#)}}}. Scores showing improved robustness are highlighted in green, including those below 1 before rounding to 2 decimal places. The corresponding absolute corruption errors are in Table~6 of the \href{https://github.com/bgswaroop/pushpull-conv}{Supplementary Material.}}
  \tiny
  \addtolength{\tabcolsep}{-0.5pt} 
  \newcommand{\gr}[0]{\cellcolor{green!25}}
  \centering
  \begin{tabular*}{\textwidth}{@{\extracolsep{\fill}}l l c c c c c c c c c c c c c c c c c c}
    \toprule
    \multirow{2}{*}{D} & 
    \multirow{2}{*}{Model} & 
    \multirow{2}{*}{$E$} & 
    \multirow{2}{*}{\rotatebox[origin=c]{45}{$mCE$}} & 
    \multirow{2}{*}{\rotatebox[origin=c]{45}{$mrCE$}} & 
    \multicolumn{3}{c}{\bfseries Noise} &
    \multicolumn{4}{c}{\bfseries Blur} &
    \multicolumn{3}{c}{\bfseries Weather} &
    \multicolumn{5}{c}{\bfseries Digital} \\
        \cmidrule(lr){6-8}
        \cmidrule(lr){9-12}
        \cmidrule(lr){13-15}
        \cmidrule(lr){16-20}
        &&&&& Ga & Sh & Im & De & Gl & Mo & Zo & Sn & Fr & Fo & Br & Co & El & Pi & Jp\\
    \cmidrule(lr){1-20}
    
    \multirow{6}{*}{\rotatebox[origin=c]{90}{\cifar}} 
		&    ResNet18 {\scalebox{.9}{\textcolor{teal}{(b1)}}}&     .069& .2853 &     1.00&     1.00&     1.00&     1.00&     1.00&     1.00&     1.00&     1.00&     1.00&     1.00&     1.00&     1.00&     1.00&     1.00&     1.00&     1.00 \\
		&     PushPull avg3&     .071& .2728 & \gr .970& \gr 0.86& \gr 0.86&     1.01&     1.00&     1.00& \gr 0.97&     1.01& \gr 0.98& \gr 0.90&     1.07&     1.01&     1.03& \gr 0.99& \gr 0.94& \gr 0.90 \\
		&     PushPull avg5&     .071& .2708 & \gr .965& \gr 0.88& \gr 0.89& \gr 0.98& \gr 0.99& \gr 0.92& \gr 0.96& \gr 1.00& \gr 0.98& \gr 0.96& \gr 0.99&     1.04&     1.03& \gr 1.00& \gr 0.94& \gr 0.93 \\\cmidrule(lr){2-20}
		&    ResNet50 {\scalebox{.9}{\textcolor{teal}{(b2)}}}&     .069& .3028 &     1.00&     1.00&     1.00&     1.00&     1.00&     1.00&     1.00&     1.00&     1.00&     1.00&     1.00&     1.00&     1.00&     1.00&     1.00&     1.00 \\
		&     PushPull avg3&     .076& .2909 & \gr .982& \gr 0.88& \gr 0.86& \gr 0.99& \gr 0.99& \gr 0.93&     1.04& \gr 0.97&     1.01& \gr 0.91&     1.06&     1.07&     1.09& \gr 0.99& \gr 0.98& \gr 0.94 \\
		&     PushPull avg5&     .068& .2866 & \gr .959& \gr 0.89& \gr 0.89& \gr 0.94& \gr 0.97& \gr 0.95& \gr 1.00& \gr 0.94& \gr 0.97& \gr 0.90& \gr 0.98&     1.02&     1.06& \gr 0.97& \gr 0.97& \gr 0.95 \\
        \cmidrule(lr){1-20}
        
    \multirow{6}{*}{\rotatebox[origin=c]{90}{\textsc{ImgNet100}}} 
		&    ResNet18 {\scalebox{.9}{\textcolor{teal}{(b3)}}}&     .207& .5812 &     1.00&     1.00&     1.00&     1.00&     1.00&     1.00&     1.00&     1.00&     1.00&     1.00&     1.00&     1.00&     1.00&     1.00&     1.00&     1.00 \\
		&     PushPull avg3&     .208& .5594 & \gr .961& \gr 0.92& \gr 0.92& \gr 0.92&     1.00& \gr 0.95& \gr 0.99&     1.01&     1.01& \gr 0.99& \gr 0.97&     1.01&     1.00& \gr 0.97& \gr 0.79& \gr 0.96 \\
		&     PushPull avg5&     .221& .5662 & \gr .976& \gr 0.92& \gr 0.91& \gr 0.92&     1.02& \gr 0.94& \gr 1.00& \gr 1.00&     1.02&     1.01&     1.02&     1.06&     1.02&     1.00& \gr 0.85& \gr 0.95 \\\cmidrule(lr){2-20}
		&    ResNet50 {\scalebox{.9}{\textcolor{teal}{(b4)}}}&     .189& .5691 &     1.00&     1.00&     1.00&     1.00&     1.00&     1.00&     1.00&     1.00&     1.00&     1.00&     1.00&     1.00&     1.00&     1.00&     1.00&     1.00 \\
		&     PushPull avg3&     .194& .5517 & \gr .972& \gr 0.95& \gr 0.95& \gr 0.95& \gr 0.97& \gr 0.91&     1.01&     1.00& \gr 0.96& \gr 0.98&     1.00&     1.04& \gr 0.99&     1.00& \gr 0.90& \gr 0.96 \\
		&     PushPull avg5&     .203& .5452 & \gr .962& \gr 0.91& \gr 0.92& \gr 0.91& \gr 0.95& \gr 0.88& \gr 0.97& \gr 0.95& \gr 0.99& \gr 0.98&     1.05&     1.06&     1.02& \gr 0.99& \gr 0.92& \gr 0.93 \\
        \cmidrule(lr){1-20}
    \multirow{6}{*}{\rotatebox[origin=c]{90}{\textsc{ImgNet200}}} 
		&    ResNet18 {\scalebox{.9}{\textcolor{teal}{(b5)}}}&     .246& .6424 &     1.00&     1.00&     1.00&     1.00&     1.00&     1.00&     1.00&     1.00&     1.00&     1.00&     1.00&     1.00&     1.00&     1.00&     1.00&     1.00 \\
		&     PushPull avg3&     .252& .6291 & \gr .979& \gr 0.97& \gr 0.97& \gr 0.96&     1.00& \gr 0.97& \gr 0.98& \gr 0.98&     1.01& \gr 0.99&     1.03&     1.02&     1.02& \gr 0.98& \gr 0.88& \gr 0.93 \\
		&     PushPull avg5&     .253& .6280 & \gr .976& \gr 0.97& \gr 0.97& \gr 0.96& \gr 1.00& \gr 0.97& \gr 0.98& \gr 0.99& \gr 0.99& \gr 0.99&     1.02&     1.01&     1.02& \gr 0.99& \gr 0.87& \gr 0.92 \\\cmidrule(lr){2-20}
		&    ResNet50 {\scalebox{.9}{\textcolor{teal}{(b6)}}}&     .213& .6183 &     1.00&     1.00&     1.00&     1.00&     1.00&     1.00&     1.00&     1.00&     1.00&     1.00&     1.00&     1.00&     1.00&     1.00&     1.00&     1.00 \\
		&     PushPull avg3&     .217& .6076 & \gr .979& \gr 0.99& \gr 0.98& \gr 0.99&     1.01& \gr 0.97& \gr 0.98& \gr 0.99&     1.01& \gr 0.98&     1.01&     1.01& \gr 1.00& \gr 0.98& \gr 0.83& \gr 0.96 \\
		&     PushPull avg5&     .222& .6062 & \gr .978& \gr 0.98& \gr 0.97& \gr 0.98& \gr 0.99& \gr 0.96& \gr 0.98& \gr 0.99&     1.02& \gr 0.99&     1.05&     1.03&     1.03& \gr 0.98& \gr 0.84& \gr 0.90 \\
        \cmidrule(lr){1-20}
    \multirow{6}{*}{\rotatebox[origin=c]{90}{\textsc{ImageNet}}} 
		&    ResNet18 {\scalebox{.9}{\textcolor{teal}{(b7)}}}&     .325& .7167 &     1.00&     1.00&     1.00&     1.00&     1.00&     1.00&     1.00&     1.00&     1.00&     1.00&     1.00&     1.00&     1.00&     1.00&     1.00&     1.00 \\
		&     PushPull avg3&     .363& .6776 & \gr .949& \gr 0.95& \gr 0.94& \gr 0.93& \gr 0.95& \gr 0.88& \gr 0.96& \gr 0.98& \gr 0.98& \gr 0.97&     1.09&     1.08&     1.01& \gr 0.93& \gr 0.67& \gr 0.91 \\
		&     PushPull avg5&     .337& .6952 & \gr .970& \gr 0.99& \gr 0.98& \gr 0.98& \gr 0.98& \gr 0.96& \gr 0.98& \gr 0.98& \gr 0.99& \gr 0.98&     1.01&     1.01& \gr 0.99& \gr 0.94& \gr 0.80& \gr 0.96 \\\cmidrule(lr){2-20}
		&    ResNet50 {\scalebox{.9}{\textcolor{teal}{(b8)}}}&     .269& .6674 &     1.00&     1.00&     1.00&     1.00&     1.00&     1.00&     1.00&     1.00&     1.00&     1.00&     1.00&     1.00&     1.00&     1.00&     1.00&     1.00 \\
		&     PushPull avg3&     .282& .6454 & \gr .967& \gr 0.99& \gr 0.98& \gr 0.98& \gr 0.96& \gr 0.96& \gr 0.98& \gr 0.97& \gr 0.99& \gr 1.00&     1.02& \gr 0.99& \gr 0.99& \gr 0.91& \gr 0.81& \gr 0.97 \\
		&     PushPull avg5&     .276& .6452 & \gr .966& \gr 0.99& \gr 0.98& \gr 0.99& \gr 0.98& \gr 0.95& \gr 0.97& \gr 0.97& \gr 0.97& \gr 0.97& \gr 0.99& \gr 0.98& \gr 0.99& \gr 0.92& \gr 0.83& \gr 1.00 \\
    \bottomrule
  \end{tabular*}

  \label{tab:classification_results}
\end{table*}

We consider the vanilla ResNet as the baseline for each experiment and compare it against two push-pull variants. In particular, we explore the impact of the hyper-parameter \textit{average kernel size} and keep the \textit{inhibition strength} trainable. The classification results presented in \cref{tab:classification_results} indicate that the push-pull variants consistently perform better than the respective baselines in terms of robustness. Since all scores are normalized to the baseline, error values less than 1 show an improvement over the baseline's performance. It is also worth noting that the relative improvement in robustness to the baseline does not diminish with increasing dataset size. 

We conducted a preliminary grid search to determine the baseline network's hyperparameters. The best-performing one was then used to train the push-pull models, ensuring an unbiased selection of learning hyperparameters. Although we conducted experiments with two sizes of the \textit{average filter}, we cannot determine which filter size is more appropriate for \imagenet. Perhaps this hyperparameter needs to be explored for other applications and ConvNet architectures. For the models trained on \cifar, however, pooling with average filters of size $5\times5$ shows a greater benefit than with $3\times3$ kernels. 

\begin{figure}[h]
  \hspace{-0.18in}
  \centering
   \includegraphics[width=0.75\linewidth]{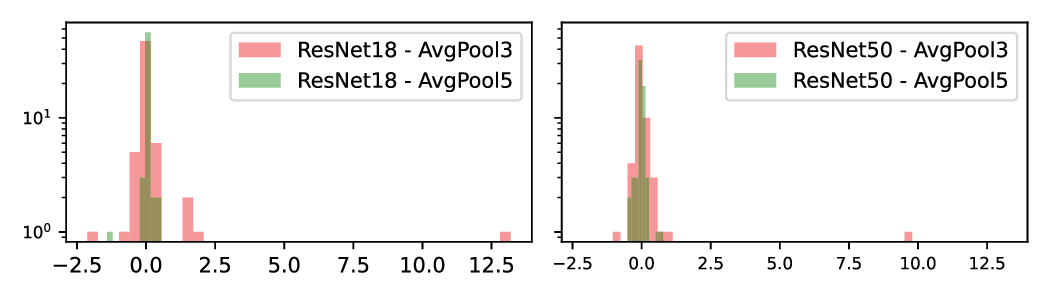}
   \caption{Distribution of learnt inhibition strength $\alpha$ for PushPull-based ResNets trained on \imagenet.}
   \label{fig:distribution_of_alpha}
\end{figure}

Another design choice in the push-pull unit is the \textit{inhibition strength} $\alpha$. For the experiments reported in \cref{tab:classification_results}, $\alpha$ was kept as a trainable parameter. \Cref{fig:distribution_of_alpha} shows the distribution of the learned $\alpha$ values. Most of the values are close to zero, indicating the network's preference to learn filters that are similar to push or with the property of band-pass filtering. With $3\times3$ average pooling, the $\alpha$ takes on values from a larger range. When $\alpha$ becomes large in magnitude, the resulting push-pull filter becomes a low-pass filter. We can also keep $\alpha$ as a fixed hyperparameter which would give us more control over the trade-off between robustness and clean accuracy, discussed further in \cref{sec: Clean error vs robustness}. 

\subsection{Comparison with the state-of-the-art}
For a fair comparison, we select methods that only alter architecture for robustness. \cite{zhang2019making} proposed using blur filters after strided convolutions to mitigate aliasing impact. \cite{vasconcelos2020effective} observed using blur filters before non-linear activations make ConvNets resilient to aliasing. We trained all models with the same training recipe. Additionally, we compare our approach with the earlier attempt at implementing the push-pull inhibition in a ConvNet~\cite{strisciuglio2020enhanced}.
All these results are summarized in \cref{tab:sota_results}, and later on discussed.  

\begin{table*}
  \caption{Comparison with the state-of-the-art. 
  BlurConv1~\cite{zhang2019making}, BlurConv2~\cite{vasconcelos2020effective}, the previous attempt of PushPull~\cite{strisciuglio2020enhanced}, and with EfficientNet~\cite{tan2019efficientnet}. The corresponding absolute corruption errors are in Table~7 of the \href{https://github.com/bgswaroop/pushpull-conv}{Supplementary Material.}}
  \tiny
  \addtolength{\tabcolsep}{-0.5pt} 
  \newcommand{\gr}[0]{\cellcolor{green!25}}
  \centering
  \begin{tabular*}{\textwidth}{@{\extracolsep{\fill}}l l c c c c c c c c c c c c c c c c c c}
    \toprule
    \multirow{2}{*}{D} & 
    \multirow{2}{*}{Model} & 
    \multirow{2}{*}{$E$} & 
    \multirow{2}{*}{\rotatebox[origin=c]{45}{$mCE$}} & 
    \multirow{2}{*}{\rotatebox[origin=c]{45}{$mrCE$}} & 
    \multicolumn{3}{c}{\bfseries Noise} &
    \multicolumn{4}{c}{\bfseries Blur} &
    \multicolumn{3}{c}{\bfseries Weather} &
    \multicolumn{5}{c}{\bfseries Digital} \\
        \cmidrule(lr){6-8}
        \cmidrule(lr){9-12}
        \cmidrule(lr){13-15}
        \cmidrule(lr){16-20}
        &&&&& Ga & Sh & Im & De & Gl & Mo & Zo & Sn & Fr & Fo & Br & Co & El & Pi & Jp\\
    \cmidrule(lr){1-20}
    
    \multirow{6}{*}{\rotatebox[origin=c]{90}{\cifar}} 
		&     ResNet50 {\scalebox{.9}{\textcolor{teal}{(b2)}}}    &   .069& .2809 &     1.00&     1.00&     1.00&     1.00&     1.00&     1.00&     1.00&     1.00&     1.00&     1.00&     1.00&     1.00&     1.00&     1.00&     1.00&     1.00 \\
		&     PushPull avg3 (ours)      &   .076& .2681 & \gr .982& \gr 0.88& \gr 0.86& \gr 0.99& \gr 0.99& \gr 0.93&     1.04& \gr 0.97&     1.01& \gr 0.91&     1.06&     1.07&     1.09& \gr 0.99& \gr 0.98& \gr 0.94 \\
            &     PushPull avg5 (ours)      &   .068& .2808 & \gr .959& \gr 0.89& \gr 0.89& \gr 0.94& \gr 0.97& \gr 0.95&     1.00& \gr 0.94& \gr 0.97& \gr 0.90& \gr 0.98&     1.02&     1.06& \gr 0.97& \gr 0.97& \gr 0.95 \\
		&     PushPull in~\cite{strisciuglio2020enhanced}                 &   .069& .2984 & \gr .973& \gr 0.94& \gr 0.92& \gr 0.93& \gr 0.97& \gr 0.94&     1.01& \gr 0.97& \gr 0.99& \gr 0.91&     1.02& \gr 0.99&     1.07& \gr 0.99& \gr 0.97& \gr 0.97 \\
		&     BlurConv1                 &   .069& .2831 & \gr .948& \gr 0.85& \gr 0.86&     1.00& \gr 0.95& \gr 0.89& \gr 0.96& \gr 0.99& \gr 0.95& \gr 0.93& \gr 0.98& \gr 0.96& \gr 0.94& \gr 0.96& \gr 0.98& \gr 0.99 \\
		&     BlurConv2                 &   .098& .3648 &     1.24&     1.06&     1.10&     1.07&     1.34&     1.13&     1.46&     1.34&     1.27&     1.14&     1.35&     1.41&     1.37&     1.29&     1.26&     1.08 \\
            \cmidrule(lr){1-20}
    \multirow{17}{*}{\rotatebox[origin=c]{90}{\imagenet}} 
		&     ResNet50  {\scalebox{.9}{\textcolor{teal}{(b8)}}}   &   .269& .6670 &     1.00&     1.00&     1.00&     1.00&     1.00&     1.00&     1.00&     1.00&     1.00&     1.00&     1.00&     1.00&     1.00&     1.00&     1.00&     1.00 \\
		&     PushPull avg3 (PP3)       &   .282& .6450 & \gr .967& \gr 0.99& \gr 0.98& \gr 0.98& \gr 0.96& \gr 0.96& \gr 0.98& \gr 0.97& \gr 0.99&     1.00&     1.02& \gr 0.99& \gr 0.99& \gr 0.91& \gr 0.81& \gr 0.97 \\
            &     PushPull avg5 (PP5)       &   .276& .6450 & \gr .966& \gr 0.99& \gr 0.98& \gr 0.99& \gr 0.98& \gr 0.95& \gr 0.97& \gr 0.97& \gr 0.97& \gr 0.97& \gr 0.99& \gr 0.98& \gr 0.99& \gr 0.92& \gr 0.83&     1.00 \\
      	&     PushPull in~\cite{strisciuglio2020enhanced}                &   .269& .6650 & \gr .996&     1.00&     1.00&     1.02& \gr 0.99&     1.00& \gr 0.99& \gr 0.99& \gr 0.99& \gr 0.98&     1.00& \gr 0.98& \gr 0.99& \gr 0.99& \gr 0.99&     1.03 \\
		&     BlurConv1                 &   .265& .6260 & \gr .962& \gr 0.97& \gr 0.96& \gr 0.97& \gr 0.97& \gr 0.97& \gr 0.98& \gr 0.97& \gr 0.97& \gr 0.96& \gr 0.95& \gr 0.96& \gr 0.97& \gr 0.95& \gr 0.94& \gr 0.95 \\
            &     BlurConv2                 &   .276& .6210 & \gr .938& \gr 0.90& \gr 0.89& \gr 0.89& \gr 0.99& \gr 0.97& \gr 0.99& \gr 0.96&     1.00& \gr 0.96& \gr 0.98&     1.00&     1.00& \gr 0.98& \gr 0.70& \gr 0.87 \\
            &     BlurConv2+PP3           &   .307& .6166 & \gr .924& \gr 0.91& \gr 0.89& \gr 0.89& \gr 0.95& \gr 0.89& \gr 0.96& \gr 0.92&     1.01& \gr 0.97& \gr 1.09&     1.08&     1.03& \gr 0.91& \gr 0.55& \gr 0.87 \\\cmidrule(lr){2-20}   
            &	Res50+AugMix	{\scalebox{.9}{\textcolor{teal}{(b9)}}} &	.269	& .6121 &	1.00	&	1.00	&	1.00	&	1.00	&	1.00	&	1.00	&	1.00	&	1.00	&	1.00	&	1.00	&	1.00	&	1.00	&	1.00	&	1.00	&	1.00	&	1.00	\\
            &	Res50+AugMix+PP3	&	.287	& .5900 &	\gr	.968	&	\gr	0.95	&	\gr	0.94	&	\gr	0.93	&	\gr	0.98	&	\gr	0.94	&	\gr	0.99	&	\gr	0.95	&	1.01	&	\gr	0.99	&	1.06	&	1.02	&	1.00	&	\gr	0.89	&	\gr	0.84	&	1.02	\\\cmidrule(lr){2-20}
            &	Res50+AutoAug	{\scalebox{.9}{\textcolor{teal}{(b10)}}} &	.269	& .6298 &	1.00	&	1.00	&	1.00	&	1.00	&	1.00	&	1.00	&	1.00	&	1.00	&	1.00	&	1.00	&	1.00	&	1.00	&	1.00	&	1.00	&	1.00	&	1.00	\\
            &	Res50+AutoAug+PP3	&	.283	& .6041 &	\gr	.961	&	\gr	0.95	&	\gr	0.95	&	\gr	0.94	&	\gr	0.97	&	\gr	0.97	&	\gr	0.99	&	\gr	0.98	&	1.00	&	\gr	0.98	&	1.02	&	1.01	&	\gr	0.95	&	\gr	0.93	&	\gr	0.82	&	0.96	\\\cmidrule(lr){2-20}
            &	Res50+PRIME	{\scalebox{.9}{\textcolor{teal}{(b11)}}} &	.298	& .5223 &	1.00	&	1.00	&	1.00	&	1.00	&	1.00	&	1.00	&	1.00	&	1.00	&	1.00	&	1.00	&	1.00	&	1.00	&	1.00	&	1.00	&	1.00	&	1.00	\\
            &	Res50+PRIME+PP3	&	.306	& .4995 &	\gr	.954	&	\gr	0.91	&	\gr	0.90	&	\gr	0.89	&	\gr	0.99	&	\gr	0.97	&	\gr 1.00	&	1.01	&	\gr	0.99	&	\gr	0.98	&	1.00	& \gr	1.00	&	\gr	0.96	&	\gr	0.96	&	\gr	0.84	&	0.90	\\\cmidrule(lr){2-20}
            &	Res50+TrivialAug	{\scalebox{.9}{\textcolor{teal}{(b12)}}} &	.269	& .5961 &	1.00	&	1.00	&	1.00	&	1.00	&	1.00	&	1.00	&	1.00	&	1.00	&	1.00	&	1.00	&	1.00	&	1.00	&	1.00	&	1.00	&	1.00	&	1.00	\\
            &	Res50+TrivialAug+PP3	&	.286	& .5867 &	\gr	.985	&	1.02	&	1.02	&	\gr	0.99	&	1.00	&	\gr	0.96	&	1.01	&	\gr	0.99	& \gr	1.00	&	1.00	&	1.07	&	1.03	&	\gr	0.96	&	\gr	0.92	&	\gr	0.83	&	0.97	\\\cmidrule(lr){2-20}
            &	EffNet-b0 {\scalebox{.9}{\textcolor{teal}{(b13)}}} &	.286	& .6818 &	1.00	&	1.00	&	1.00	&	1.00	&	1.00	&	1.00	&	1.00	&	1.00	&	1.00	&	1.00	&	1.00	&	1.00	&	1.00	&	1.00	&	1.00	&	1.00	\\
            &	EffNet-b0 + PP3	&	.288	& .6660 &	\gr	.976	&\gr	0.98	&\gr	0.99	&\gr	0.97	&\gr	0.99	&\gr	0.98	&\gr	1.00	&	1.00	&\gr	1.00	&\gr 0.99 & 1.00 &\gr	0.99	&	1.01	&\gr	0.98	&\gr	0.91	&\gr	0.86 \\
            
  \bottomrule
  \end{tabular*}

  \label{tab:sota_results}
\end{table*}

The most effective standalone method on \cifar~is identified as BlurConv1~\cite{zhang2019making}, while on \imagenet, BlurConv2~\cite{vasconcelos2020effective} leads. Our method not only outperforms earlier approaches to push-pull mechanisms but also approaches the effectiveness of BlurConv. Notably, our method enhances robustness by altering only one aspect of the ConvNet, unlike BlurConv, which modifies all layers with strided convolutions and pooling. Hence, our method achieves robustness with minimal computational overhead. 
Furthermore, our method enhances the robustness provided by BlurConv2 when coupled together, albeit with a reduction in clean accuracy. 
For every type of corruption where BlurConv2 is effective, our PushPull-Conv method has increased robustness, with a particularly notable enhancement against pixelate corruption.

The proposed approach complements the robustness offered by data augmentation schemes. We trained ResNet50 with data augmentation schemes including AutoAug, AugMix, PRIME, and TrivialAugment. On adding the PushPull layer to these models, the robustness improved further (see \cref{tab:sota_results}). This demonstrates our method's compatibility and effectiveness in conjunction with current schemes. Moreover, in terms of absolute $mCE$, push-pull with PRIME augmentations achieves the best-ever $mCE$ score of 49.95\% with ResNet50. This is a new benchmark when considering models with about 25.5M parameters. 

Finally, the proposed method was evaluated with the EfficientNet-b0~\cite{tan2019efficientnet} architecture. We integrated the push-pull layer in the first convolutional layer, that is, in the stem of EfficientNet. 
The results presented in \cref{tab:sota_results} clearly demonstrate the robustness improvements achieved with PushPull-Conv. While further investigation into various ConvNet-based architectures is necessary, the evidence from ResNets and EfficientNet implementations shows promising outcomes. In fact, with EfficientNet we achieve robustness with negligible cost to clean error. We present a metric to quantify such a trade-off in the following section.

\section{Discussion}

\subsection{Clean error vs robustness}
\label{sec: Clean error vs robustness}

\begin{figure}
  \centering
   \includegraphics[width=0.7\linewidth]{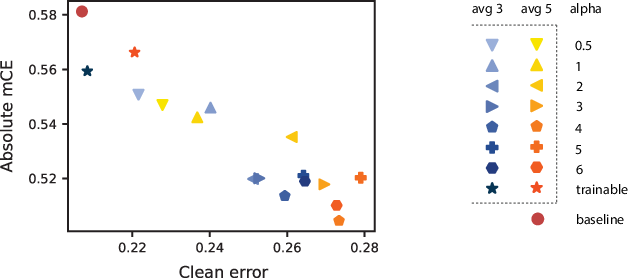}
   \caption{Comparison of PushPull-Conv configurations with vanilla ResNet18. The PushPull variants improve mean corruption error (mCE) over the baseline on \imagenetc, with adjustable hyper-parameters to control the trade-off between robustness and clean error. (Best viewed in colour)}
   \label{fig:figure_resnet18_imagenet100_classification}
\end{figure}

\noindent Tsipras~et.~al~\cite{tsipras2018robustness} explained the trade-off between robustness and generalization (clean accuracy) and concluded that the interplay between robustness and standard accuracy might be more nuanced than one might expect. We investigate this tradeoff with PushPull-Conv for various settings of hyperparameters. The inhibition strength $\alpha$ plays a key role and can be used as a knob. As depicted in \cref{fig:figure_resnet18_imagenet100_classification}, with increasing values of $\alpha$ the models become progressively resilient to image corruption while taking a dip in performance on clean data. We propose a metric to quantify the trade-off between clean error and robustness (see \cref{sec: tradeoff_metric}). The best trade-off is achieved when $\alpha$ is set as a trainable parameter and AvgPool $3\times3$.

\begin{table}[t]
  \caption{Summary of experiments using various push-pull configurations with ResNet18 trained on \imagenetonehunderd. The results of configurations labelled $\text{PushPull}^{*}$ are averaged over experiments with values of $\alpha \in \{0.5, 1, 2, 3, 4, 5, 6\}$.}
  \centering
  \scriptsize
  \begin{tabular*}{0.55\textwidth}{@{\extracolsep{\fill}}l l l}
    \toprule
    \textbf{Configuration} & \textbf{Clean error} & $\mathbf{mrCE}$ \\
    \midrule
    ResNet18 {\scalebox{.9}{\textcolor{teal}{(b3)}}}	            &$0.207$	&$1.000$ \\
    $\text{PushPull}^{*}$: no pooling	&$0.249 \scriptstyle~\pm~0.012$	&$0.932 \scriptstyle~\pm~0.012$ \\
    $\text{PushPull}^{*}$: avg 3		&$0.251 \scriptstyle~\pm~0.015$	&$0.929 \scriptstyle~\pm~0.015$ \\
    $\text{PushPull}^{*}$: avg 5		&$0.260 \scriptstyle~\pm~0.020$	&$0.928 \scriptstyle~\pm~0.017$ \\
    \bottomrule
  \end{tabular*}
  \label{tab:resnet18_fixed_inh_summary}
\end{table}

We evaluate the contribution of the average pooling to the overall robustness. To this end, we conducted three sets of experiments. The first set had no pooling and the other two were conducted with average pooling of sizes $3 \times 3$ and $5 \times 5$. The results are summarized in \cref{tab:resnet18_fixed_inh_summary}. 
It can be seen that the robustness improves with higher levels of pooling while it has the opposite effect on the clean error.

\subsection{A Quantitative Metric for Assessing the Trade-off Between Robustness and Clean Accuracy}
\label{sec: tradeoff_metric}
It is important to quantify the trade-off between clean error and robustness. \Cref{fig:figure_resnet18_imagenet100_classification} illustrates this trade-off when considering ResNet18 models trained with several configurations of hyperparameters on \imagenetonehunderd. To identify which configuration of hyperparameters results in the best trade-off, we need to quantify it. Therefore, we introduce a metric, which quantifies the net reduction in error rate of a model with respect to a baseline. 

For a given model $m$, let the clean error be denoted by $E$ and the mean corruption error as $mCE$. The corresponding errors for the baseline are denoted by $E_{0}$ and $mCE_{0}$. On considering clean error, the net reduction in error rate for a model $m$ w.r.t a baseline is given by:

\begin{equation}
    R_E = \frac{E_0 - E}{E_0}
\end{equation}

\noindent Likewise, the reduction in error rate for $mCE$ is given by:

\begin{equation}
    R_{mCE} = \frac{mCE_0 - mCE}{mCE_0}
\end{equation}

\noindent The net reduction in error rate is then determined by:

\begin{equation}
    R_{net} = \frac{1}{2}(R_E + \beta \times R_{mCE})
\end{equation}

\noindent where the factor $\beta$ determines the difficulty in reducing $mCE$ over $E$.  
When $\beta=mCE_{0}/E_{0}$, it quantifies the difficulty faced by the baseline to achieve the same level of robustness as the clean accuracy.
Setting $\beta=1$ implies giving equal weight to both robustness and accuracy, without accounting for the difficulty of the problem. For a given model $m$, when $R_{net}=0$ it implies the model is equivalent in terms of net performance to the baseline. When $R_{net}>0$, the performance tradeoff favours model $m$ compared to the baseline. Conversely, $R_{net}<0$ indicates a negative tradeoff with respect to the baseline. 

\begin{figure}
  \centering
   \includegraphics[width=0.9\linewidth]{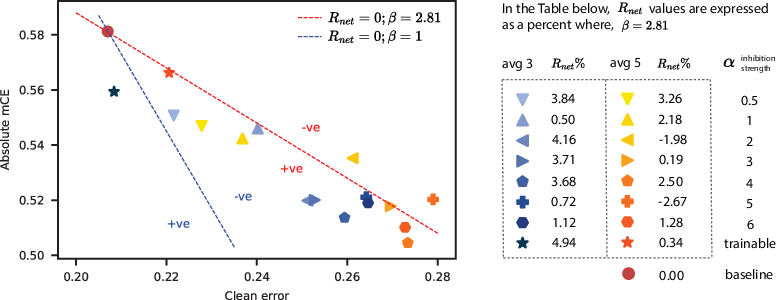}
   \caption{Comparison of PushPull-Conv configurations with baseline vanilla ResNet18 trained on \imagenetonehunderd. The dotted lines indicate the points at which the models do not make any improvements over the baseline in terms of net reduction in error rate. Each line splits the error space into two regions ($+$ve and $-$ve). The models in the $+$ve regions are the favourable ones with $R_{net}>0$ and vice-versa. (Best viewed in colour)}
   \label{fig: tradeoff metric}
\end{figure}

\Cref{fig: tradeoff metric} illustrates the robustness of the push-pull variants compared to the baseline. It is interesting to note that setting $\beta=mCE_{0}/E_{0}=2.81$, accounts for the difficulty of improving robustness over clean accuracy. In this scenario, almost all variants of push-pull positively reduce the error over the baseline. Even when considering equal difficulty for both problems ($\beta=1$), the push-pull variant with AvgPool $3\times3$ and trainable inhibition performs better than the baseline. 

\subsection{Robustness to high-frequency corruptions}
In most experiments, a trend was observed where the PushPull models achieved higher robustness to high- and mid-frequency corruptions compared to their low-frequency counterparts. To illustrate this, we categorized the 15 corruption types from the \cifar-C dataset into low-, mid-, or high-frequency corruptions based on the analysis of~\cite{yin2019fourier}. Frost, fog, brightness, and contrast are classified as low-frequency corruptions; glass-blur, motion-blur, zoom, snow, and elastic are classified into mid-frequency corruptions, and the rest as high-frequency. \Cref{tab:mrCE per frequency} shows the $mrCE$ scores when the corruption types are grouped based on their frequency. The results indicate that PushPull offers more robustness to high-frequency corruptions than low frequencies. This trend is consistent with the other datasets and ResNet models.

\begin{table}
\caption{The averaged $mrCE$ scores for \cifar-C corruption types, grouped by corruption frequency. All PushPull models are trained with trainable inhibition $\alpha$.}
  \centering
  \scriptsize
  \begin{tabular*}{0.60\textwidth}{@{\extracolsep{\fill}}l l l l}
    \toprule
    \textbf{Model} & \textbf{Low freq.} & \textbf{Mid freq.} & \textbf{High freq.} \\
    \midrule
	ResNet18 {\scalebox{.9}{\textcolor{teal}{(b1)}}} &  $1.00\scriptstyle~\pm~0.000$&	$1.00\scriptstyle~\pm~0.000$&	$1.00\scriptstyle~\pm~0.000$ \\
    PushPull avg3	  &	 $1.00\scriptstyle~\pm~0.074$&	$0.99\scriptstyle~\pm~0.013$&	$0.93\scriptstyle~\pm~0.069$ \\
    PushPull avg5	  &	 $1.01\scriptstyle~\pm~0.036$&	$0.97\scriptstyle~\pm~0.032$&	$0.93\scriptstyle~\pm~0.046$ \\ \cmidrule(lr){2-4}    
    ResNet50 {\scalebox{.9}{\textcolor{teal}{(b2)}}} &	 $1.00\scriptstyle~\pm~0.000$&	$1.00\scriptstyle~\pm~0.000$&	$1.00\scriptstyle~\pm~0.000$ \\
    PushPull avg3	  &	 $1.03\scriptstyle~\pm~0.025$&	$0.99\scriptstyle~\pm~0.026$&	$0.94\scriptstyle~\pm~0.032$ \\
    PushPull avg5	  &	 $0.99\scriptstyle~\pm~0.019$&	$0.97\scriptstyle~\pm~0.013$&	$0.93\scriptstyle~\pm~0.040$ \\	
    \bottomrule
  \end{tabular*}
  \label{tab:mrCE per frequency}
\end{table}

\section{Conclusion}
PushPull-based convolutions improve the robustness of ResNet architectures to out-of-distribution image corruptions, especially those with high-frequency corruptions. This robustness is achieved without data augmentation and with a relatively small computational cost. Furthermore, this approach can be combined with data augmentation. In particular, when combined with PRIME augmentation, it achieves the best-ever reported $mCE$ of 49.95\% using ResNet50 on \imagenetc.

\bibliography{bibliography}
\bibliographystyle{splncs04}

\end{document}